\documentclass[10pt,journal,compsoc]{IEEEtran}

\ifCLASSINFOpdf
\else
\fi

\usepackage{graphicx}
\usepackage{times}
\usepackage{epsfig}
\usepackage{graphicx}
\usepackage{amsmath}
\usepackage{amssymb}
\usepackage{booktabs}
\usepackage{multirow}
\graphicspath{{fig/}}
\usepackage[pagebackref=true,breaklinks=true,colorlinks,bookmarks=false]{hyperref}
\usepackage{float}

\usepackage{tabularx}

\usepackage{epstopdf}
\usepackage{times}
\usepackage{epsfig}
\usepackage{graphicx}
\usepackage{amsmath}
\usepackage{amssymb}

\usepackage{nicefrac}  
\usepackage{capt-of}
\usepackage{amsfonts} 
\usepackage{booktabs}
\usepackage{mathtools}
\usepackage{amsthm}
\usepackage{xspace}
\usepackage{amstext}
\usepackage{enumitem}
\usepackage[ruled,vlined,algo2e]{algorithm2e}
\usepackage{algorithm}
\usepackage{algorithmic}
\usepackage{multirow}

\usepackage{wrapfig}
\usepackage[accsupp]{axessibility}

\newcommand{\bfsection}[1]{\vspace*{0.1cm}\noindent\textbf{#1}}
\newtheorem{prop}{Proposition}

\def\wrt{\emph{w.r.t. }}
\DeclareMathOperator{\arccosh}{arcosh}

\usepackage{newpxtext}

\makeatletter
\newcommand{\ie}{\emph{i.e.}\@ifnextchar.{\!\@gobble}{}}
\newcommand{\eg}{\emph{e.g.}\@ifnextchar.{\!\@gobble}{}}
\newcommand{\etc}{etc\@ifnextchar.{}{.\@}}
\newcommand{\etal}{\textit{et al.}}
\makeatother

\begin{document}
%


\title{Hyperbolic Chamfer Distance for Point Cloud Completion and Beyond}

%
%
%
%

\author{Fangzhou~Lin, Songlin~Hou, Haotian~Liu,  Shang~Gao, Kazunori~D~Yamada, Haichong~K.~Zhang~\IEEEmembership{Member,~IEEE, Ziming~Zhang}

\IEEEcompsocitemizethanks{

\IEEEcompsocthanksitem Fangzhou Lin, Songlin Hou, Haotian Liu, Shang Gao, Dr Haichong Zhang and Dr. Ziming Zhang are with Worcester Polytechnic Institute, Worcester, MA 01609, USA. Fangzhou Lin, Haotian Liu, Shang Gao and Dr Haichong Zhang are from Robotics Engineering. Songlin Hou and Dr. Ziming Zhang are from  Electrical \& Computer
Engineering. Songlin Hou is also with Dell Technologies, Hopkinton, MA 01748,  USA.  E-mail: \{flin2,  shou, hliu8, sgao, zzhang15\}@wpi.edu

\IEEEcompsocthanksitem Dr. Kazunori D Yamada is with Unprecedented-scale Data Analytics Center and Graduate School of Information Sciences of Tohoku University, Sendai, Sendai, 9800845, Japan. E-mail: yamada@tohoku.ac.jp}

\thanks{Dr. Haichong K Zhang and Dr. Ziming Zhang are the corresponding authors. Source code is available at \url{https://github.com/Zhang-VISLab}.}

}

\markboth{Journal of \LaTeX\ Class Files,~Vol.~14, No.~8, August~2015}%
{Shell \MakeLowercase{\textit{et al.}}: Bare Demo of IEEEtran.cls for IEEE Journals}
%



\IEEEtitleabstractindextext{%

\begin{abstract}
Chamfer Distance (CD) is widely used as a metric to quantify difference between two point clouds. In point cloud completion, Chamfer Distance (CD) is typically used as a loss function in deep learning frameworks. However, it is generally acknowledged within the field that Chamfer Distance (CD) is vulnerable to the presence of outliers, which can consequently lead to the convergence on suboptimal models. In divergence from the existing literature, which largely concentrates on resolving such concerns in the realm of Euclidean space, we put forth a notably uncomplicated yet potent metric specifically designed for point cloud completion tasks: {Hyperbolic Chamfer Distance (HyperCD)}. This metric conducts Chamfer Distance computations within the parameters of hyperbolic space. During the backpropagation process, HyperCD systematically allocates greater weight to matched point pairs exhibiting reduced Euclidean distances. This mechanism facilitates the preservation of accurate point pair matches while permitting the incremental adjustment of suboptimal matches, thereby contributing to enhanced point cloud completion outcomes. Moreover, measure the shape dissimilarity is not solely work for point cloud completion task, we further explore its applications in other generative related tasks, including single image reconstruction from point cloud, and upsampling. We demonstrate state-of-the-art performance on the point cloud completion benchmark datasets, \ie PCN, ShapeNet-55, and ShapeNet-34, and show from visualization that HyperCD can significantly improve the surface smoothness, we also provide the provide experimental results beyond completion task. 
\end{abstract}

}

\maketitle

%
\IEEEdisplaynontitleabstractindextext
\IEEEpeerreviewmaketitle

\IEEEraisesectionheading{\section{Introduction}\label{sec:introduction}}


\IEEEPARstart{P}{oint} clouds, easily obtainable yet fundamentally important, are pivotal in shaping today's landscape of robotics and automation~\cite{wang2022pointattn,ma2022completing,shi2022temporal}, with applications ranging from scene modeling~\cite{peng2022epar}, field measurement~\cite{huang2018trunk} to industrial designs~\cite{tong2015low}. However, raw data of point clouds captured by existing 3D sensors is usually incomplete and sparse due to factors such as occlusion, limited sensor resolution and light reflection~\cite{yu2018pu,li2021point,luo2021score,li2021high,zhou2022seedformer}, which may adversely affect the performance of downstream tasks requiring high-quality representation, such as point cloud segmentation and detection. Researchers aim to resolve this challenge by reconstructing the full shape of an object or scene from incomplete raw point clouds. This task is referred to as {\em point cloud completion} \cite{alliegro2021denoise}. During this effort, accurately determining the similarity of two collections of points becomes a crucial part of this task. 

\begin{figure}[t]
    \hfill
	\begin{minipage}[b]{0.490\columnwidth}
		\begin{center}
			\centerline{\includegraphics[width=.9\linewidth, keepaspectratio,]{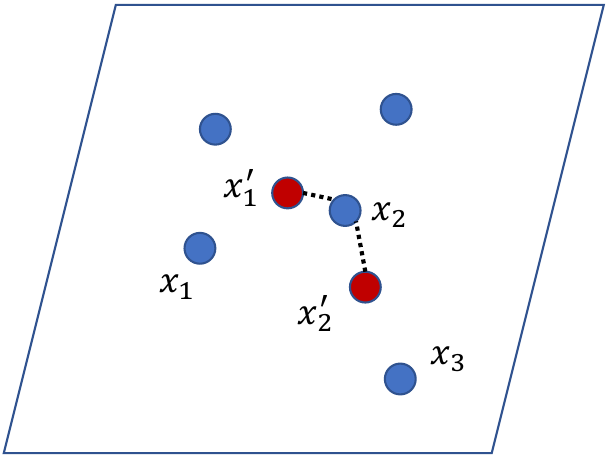}}
			\centerline{\small (a) Euclidean}
		\end{center}
	\end{minipage}
	\hfill
	\begin{minipage}[b]{0.495\columnwidth}
		\begin{center}
			\centerline{\includegraphics[width=.9\linewidth,keepaspectratio]{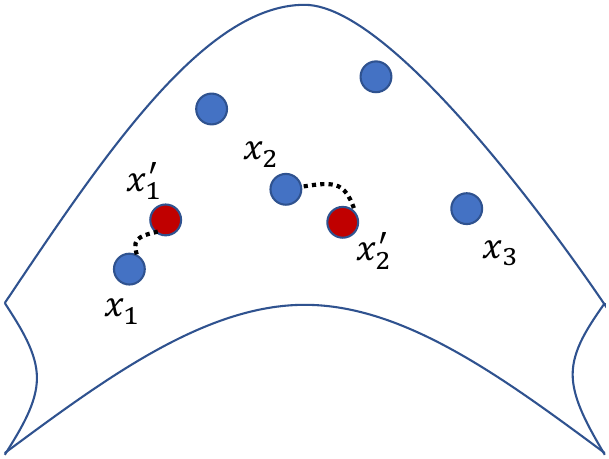}}
			\centerline{\small (b) Hyperbolic}
		\end{center}
	\end{minipage}
    \caption{Illustration of point matching in the {\bf (a)} Euclidean space and {\bf (b)} hyperbolic space. With the {\em position-aware} embeddings in hyperbolic space, the mismatched point pairs in Euclidean space may be corrected, leading to better completion performance.}
\label{fig:idea}
\end{figure}

Point cloud completion has long been regarded as a complex task due to the unordered and unstructured nature of point clouds (particularly those acquired from real-world environments).
Recently, a variety of deep learning-based approaches have been introduced for point cloud completion, including supervised learning, self-supervised learning, and unsupervised learning\cite{pcn, cascade,mittal2021self,cai2022learning,fan2022reconstruction,ren2022self}. Supervised learning, utilizing a general encoder-decoder architecture, has emerged as the predominant paradigm among researchers, delivering state-of-the-art performance across nearly all major benchmarks. Many related research centers on developing various structures within the encoder and decoder to enhance feature extraction and improve point cloud generation \cite{pointtr, snowflakenet, zhou2022seedformer, wang2022pointattn,fei2022comprehensive}, in Euclidean space.



\bfsection{Unequal Point Importance in Point Clouds.}
The visual quality of point clouds is typically perceived by humans in a non-homogeneous manner, with greater emphasis placed on points featuring specific geometric structures, such as planes, edges, and corners, \etc. For example, point clouds featuring smooth surfaces and sharp edges are generally considered more visually appealing compared to those lacking these characteristics \cite{xu2022fpcc,liu2022integrated}. This seemingly simple, yet nontrivial fact in point clouds, however, remains largely unexplored in the literature of point cloud completion. 
Chamfer distance (CD), as demonstrated in works such as \cite{guo2020deep, wu2021densityaware}, is commonly employed in point cloud completion to quantify the dissimilarity in shape between two point clouds. This is achieved by computing the average distance from each point in one set to its nearest neighbor in the other set. As much as CD accurately represents the global dissimilarity between the prediction and ground truth, it assigns equal importance to the distances of all nearest-neighbor pairs from both sets, including outliers, which results in an increased sensitivity to outliers. 

\begin{figure*}[t]
    \hfill
	\begin{minipage}[b]{\columnwidth}
		\begin{center}
			\centerline{\includegraphics[width=\linewidth, keepaspectratio,]{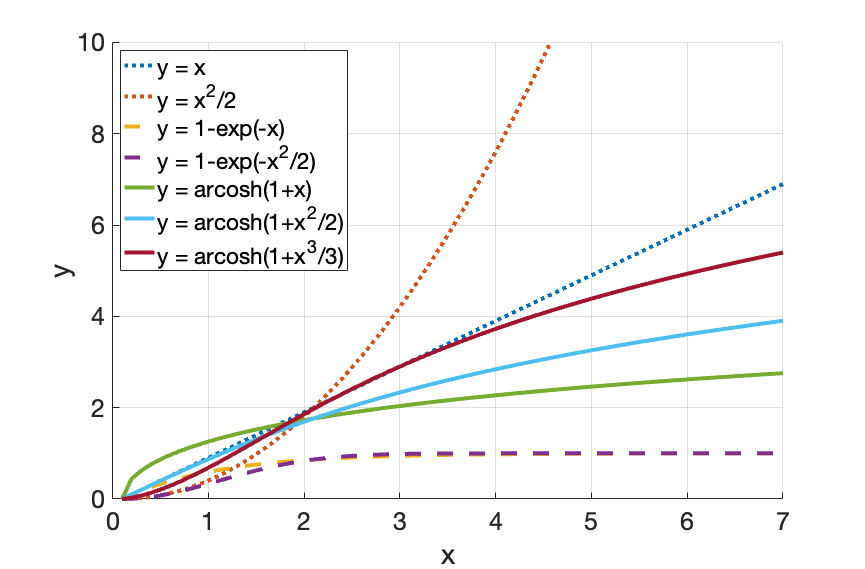}}
		\end{center}
	\end{minipage}
	\hfill
	\begin{minipage}[b]{\columnwidth}
		\begin{center}
			\centerline{\includegraphics[width=\linewidth,keepaspectratio]{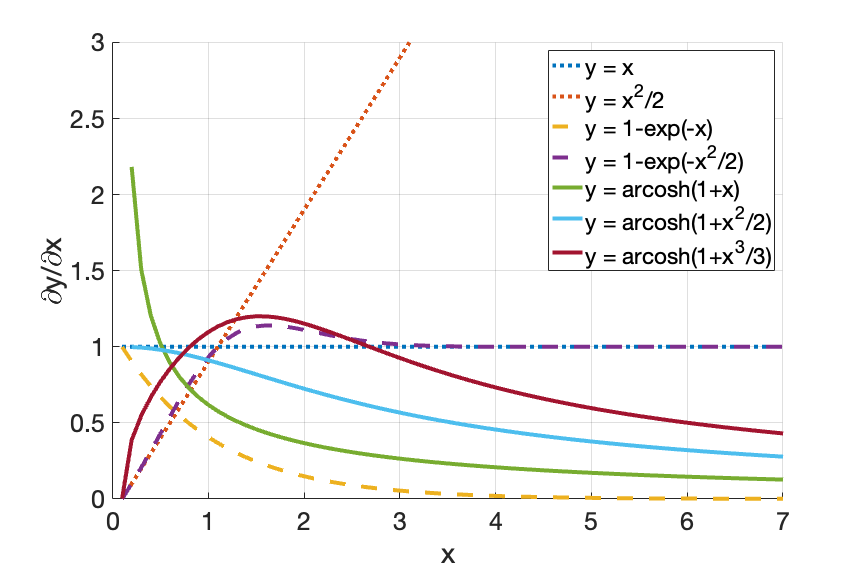}}
		\end{center}
	\end{minipage}
	\vspace{-3mm}
    \caption{Illustration of {\bf (left)} some distance metrics and {\bf (right)} their corresponding gradients, where the dotted curves are used in $\ell_1$ and $\ell_2$-CD, the dash ones are used in density-aware Chamfer distance (DCD) \cite{wu2021densityaware}, and the solid curves are special cases of our HyperCD.} 
\label{fig:function}
\end{figure*}

\bfsection{Density-aware Chamfer Distance (DCD) in Euclidean Space.}
To tackle the issue of equal weighting in CD, Wu \etal \cite{wu2021densityaware} recently introduced a DCD metric that examines the disparity in density distributions within point clouds. As shown in Fig. \ref{fig:idea} (a), denser points in point clouds are more likely to have multiple matches, whereas sparser points may have fewer matches. The DCD metric incorporates this disparity by applying a weighting mechanism (inverse to the number of matches) to ensure that sparser points are assigned higher weights. Additionally, DCD employs an exponential approximation, based on the first-order Taylor expansion, to reduce sensitivity to outliers, as demonstrated in Fig. \ref{fig:function}.

While empirical evidence suggests that DCD outperforms CD in point cloud completion, it might exhibit the following significant challenges:
\begin{itemize}[nosep, leftmargin=*]
    \item {\em Density-aware mechanism in DCD may assign higher weights to sparser points.} It not only helps achieve accurate matches at the edges and corners, but also increases the likelihood of matching outliers, resulting in a suboptimal completion.

    \item {\em CD approximate functions hardly preserve good matches.} From Fig. \ref{fig:function}, CD is susceptible to outliers due to its gradient, which assigns greater (or equal) weights to points located farther away. While DCD addresses this issue, it encounters challenges where the weights either decay too rapidly (exponentially) with the $\ell_1$ distance or become excessively small for accurate matches (even zero for perfect matches) when using the $\ell_2$ distance.
\end{itemize}


\bfsection{Hyperbolic Chamfer Distance (HyperCD).}
To address the challenges in CD for point cloud completion, we propose that the key to success lies in gradually improving poor matches while maintaining strong ones throughout training. This approach, which we term "position-aware," focuses solely on point positions. Recent research has shown that hyperbolic space can effectively represent the inherent compositional nature of point clouds by using position-aware embeddings within tree-like geometric structures, as demonstrated in works such as \cite{montanaro2022rethinking}. This insight strongly motivates our exploration of CD within hyperbolic space.

As shown in Fig. \ref{fig:idea} (b), hyperbolic space offers greater flexibility than Euclidean distance for measuring distances between points, which may help address matching errors in CD. Furthermore, as depicted in Fig. \ref{fig:function} (left), various power functions in hyperbolic space (defined by $\arccosh$) provide a closer approximation to CD than DCD. Additionally, the gradient curves of $y=\arccosh(1+x)$ and $y=\arccosh(1+x^3/3)$ closely resemble those of DCD, while the gradient curve of $y=\arccosh(1+x^2/2)$ demonstrates a precise fit, aligning with the expected behavior for an effective weighting mechanism in point cloud completion. This observation offers valuable insights for defining HyperCD.

Specifically, by matching the points with nearest neighbors in Euclidean space (represented by $x$-axis in Fig. \ref{fig:function}), we first obtain the matches between the prediction and ground truth, and vice versa. These Euclidean distances are subsequently input into $\arccosh$ to map them into hyperbolic space. Through empirical analysis, we show that models trained with this straightforward metric can significantly outperform those trained with CD and DCD.

\bfsection{Contributions.}
We list our main contributions as follows:
\begin{itemize}[nosep, leftmargin=*]
    \item We introduce HyperCD, a highly effective and straightforward distance metric for point cloud completion. To the best of our knowledge, this is the first exploration of hyperbolic space in the context of point cloud completion.

    \item We demonstrate our approach achieves state-of-the-art performance on various benchmark datasets, utilizing widely recognized networks trained with HyperCD

    \item  expand the application of HypperCD to point cloud self-supervision pre-training, single image reconstruction, and upsampling, demonstrating significant improvements over their vanilla CD metric baseline.

\end{itemize}

\section{Related Work}
\label{sec:related_work}
\bfsection{3D Shape Completion.} Previous approaches to 3D shape completion predominantly concentrate on voxel grids, utilizing network architectures that closely resemble those employed in 2D image processing \cite{maturana2015voxnet,dai2017shape,le2018pointgrid}.
When intermediate representations are utilized, information loss is unavoidable, while voxelization leads to significant computational overhead relative to voxel resolution \cite{wang2021voxel}. As a result, advanced models have been developed to process raw point cloud data directly. Building on the foundational work of PointNet \cite{qi2017pointnet}, which independently applies MLPs to each point and utilizes max-pooling to aggregate features for permutation invariance, a natural approach is to leverage permutation-invariant neural networks for designing an encoder to extract partial input features and a decoder for completing point clouds. As the first learning-based network for point cloud completion, PCN \cite{pcn} extracts global features in a manner akin to PointNet and employs folding operations \cite{yang2018foldingnet} for point generation. To capture local structures among points, Zhang \etal \cite{zhang2020detail} enhance performance by extracting multi-scale features across various layers in the feature extraction stage. Additionally, CDN \cite{wang2020cascaded} introduces a cascaded refinement network that integrates the local details of partial inputs with global shape information.

Lyu \etal \cite{lyu2021conditional} approached point cloud completion by framing it as a conditional generation task within the denoising diffusion probabilistic model (DDPM) framework \cite{sohl2015deep,ho2020denoising,zhou20213d,luo2021diffusion}. Highlighting the insensitivity of CD loss to overall density distribution, they proposed leveraging DDPM to establish a one-to-one point-wise mapping between successive point clouds during the diffusion process, trained using a straightforward mean squared error loss function \cite{bickel2015mathematical}. Despite its utility, this method is computationally demanding and applicable only during the coarse point cloud generation stage.

The superiority of attention mechanisms, such as the Transformer \cite{vaswani2017attention}, lies in their capability to effectively capture long-range interactions, unlike the limited receptive fields of CNNs. For example, SA-Net \cite{SA-Net} employs a skip-attention mechanism to integrate local region details from the encoder with point features from the decoder, aiming to retain finer geometric information during point cloud generation. SnowflakeNet \cite{snowflakenet} and PointTr \cite{pointtr} emphasize Transformer-inspired designs in the decoder, while PointAttN \cite{wang2022pointattn} introduces a fully Transformer-based architecture. Collectively, these approaches highlight the effectiveness of Transformers in point cloud completion tasks.


\bfsection{3D Shape Generation.}

Advancements in point cloud completion, coupled with significant strides in 3D point cloud representation learning, have greatly propelled the field of point cloud generation forward. Generative tasks associated with point clouds can be categorized by their input types, encompassing point cloud auto-encoding, upsampling, single-image reconstruction, and the generation of novel point clouds.
Point cloud upsampling, as explored in \cite{yu2018pu, li2019pu, qian2021pu}, addresses the challenge of transforming sparse and non-uniformly distributed input point clouds into dense, uniformly distributed outputs faithfully representing the underlying surface. PU-GAN \cite{li2019pu}, a notable example, employs a generative adversarial network to bridge the gap between sparse inputs and dense outputs. Advancing this approach, PU-GCN \cite{qian2021pu} introduces a graph convolution network designed to encode robust local features for improved upsampling. Similarly, Dis-PU \cite{li2019pu} achieves strong performance by disentangling upsampling and refinement processes through a two-step network.  
In contrast, single image (view) reconstruction focuses on generating high-quality point clouds from a single image. Early efforts like PSGN \cite{fan2017point} and AtlasNet \cite{groueix2018papier} rely on convolutional and MLP-based architectures to generate point clouds. More recently, 3DAttriFlow \cite{wen20223d} achieves state-of-the-art results by combining an attribute flow pipeline with a deformation pipeline to assign disentangled semantic attributes to 3D point clouds.

\bfsection{Point Cloud Distance.} Distance in point clouds represents a non-negative function used to quantify dissimilarity. Due to the unordered nature of point clouds, shape-level distance is often computed using statistical measures of pairwise point-level distances guided by a specific assignment approach \cite{wu2021densityaware}. Chamfer Distance (CD) and its variants, known for their relatively low computational cost and fair design, have been widely adopted in learning-based point cloud completion methods \cite{deng20193d, lyu2021conditional,zhang2022attention,tang2022lake}. On the other hand, Earth Mover's Distance (EMD), another commonly utilized metric, determines the optimal mapping between sets by solving an optimization problem. While EMD can be more reliable in some scenarios, it is computationally expensive and applicable only to sets with equal numbers of points \cite{liu2020morphing, achlioptas2018learning}. To address these challenges, Wu \etal \cite{wu2021densityaware} introduced the Density-aware Chamfer Distance (DCD), offering a metric that partially balances CD's efficiency with EMD's computational demands.

\bfsection{Hyperbolic Learning.} Euclidean space, a natural extension of the intuitive and visually accessible three-dimensional space, is extensively utilized in machine learning due to its ease in handling distance and inner-product computations \cite{ganea2018hyperbolic,liu2019hyperbolic,klimovskaia2020poincare,peng2021hyperbolic}. Nevertheless, its embedding often falls short for complex, tree-like data structures observed in domains like Biology, Network Science, Computer Graphics, or Computer Vision, which inherently possess highly non-Euclidean latent geometries \cite{ganea2018hyperbolic, bronstein2017geometric}. This limitation has driven the exploration of deep neural networks within non-Euclidean spaces, particularly hyperbolic space—a Riemannian manifold characterized by constant negative curvature. Advances in hyperbolic geometry have recently reduced the disparity between Euclidean and hyperbolic embeddings by introducing critical neural network components such as multinomial logistic regression, fully connected layers, and recurrent neural networks \cite{ganea2018hyperbolic, shimizu2020hyperbolic}.

In contrast to Euclidean space, which exhibits polynomial volume growth relative to the radius, hyperbolic space $\mathbb{H}^n$ demonstrates exponential growth, making it well-suited for modeling data with tree-like structures. The utility of hyperbolic space has been validated in various domains, including natural language processing \cite{nickel2017poincare, nickel2018learning}, image segmentation \cite{weng2021unsupervised, atigh2022hyperbolic}, few-shot learning \cite{khrulkov2020hyperbolic}, zero-shot learning \cite{liu2020hyperbolic}, and metric learning with vision transformers \cite{ermolov2022hyperbolic}. For 3D object point clouds, the data inherently possess a hierarchical property, where simpler components combine to form increasingly complex structures. Recent work by \cite{montanaro2022rethinking} has demonstrated that embedding features from a point cloud classifier into hyperbolic space results in state-of-the-art supervised models for point cloud classification. Notably, points near the hyperbolic space boundary are distributed more sparsely than those closer to its center. 
Although the hierarchical relationship between part and whole may aid in object classification, it is not directly applicable to generation tasks such as point cloud completion. Conversely, the exponential property of hyperbolic embedding provides inspiration for designing a novel loss function tailored to point cloud completion tasks, with an emphasis on surface representation. As far as we know, this marks the first application of hyperbolic space in the context of point cloud completion.

\section{Method}
\label{sec:motivations}
In this section, we present HyperCD, highlighting its efficient computation and weighting mechanism in backpropagation.

\subsection{Chamfer Distance}
\bfsection{Notations.} 
In the training dataset, let $(x_i, y_i)$ represent the $i$-th pair of point clouds, where $x_i = \{x_{ij}\}$ denotes the incomplete input point cloud comprising 3D points $x_{ij}$ for all $j$, and $y_i = \{y_{ik}\}$ signifies the ground-truth point cloud containing points $y_{ik}$ for all $k$. Define $d(\cdot, \cdot)$ as a specific distance metric and $f$ as a neural network parameterized by $\omega$ that generates a new point cloud from an incomplete input point cloud.

\bfsection{Definition.} A Chamfer distance for point clouds, considering the above notations, is generally defined as follows:
\begin{align}\label{eqn:CD}
    & \hspace{-3mm} D(x_i, y_i) \nonumber \\ & \hspace{-3mm} = \frac{1}{|x_i|}\sum_j\min_k d(x_{ij}, y_{ik}) + \frac{1}{|y_i|}\sum_k\min_j d(x_{ij}, y_{ik}),
\end{align}

where $|\cdot|$ denotes the cardinality of a set. This definition allows for the instantiation of the distance metric across various geometric spaces, including
\begin{itemize}[nosep, leftmargin=*]
    \item \underline{\em Euclidean distance:} For point cloud completion, function $d$ is usually defined in Euclidean space, referring to 
    \begin{align}
        d(x_{ij}, y_{ik})=\left\{
        \begin{array}{ll}
            \|x_{ij} - y_{ik}\| & \mbox{as {\em L1-distance}} \\
            \|x_{ij} - y_{ik}\|^2 & \mbox{as {\em L2-distance}}
        \end{array}
        \right.
    \end{align}
    where $\|\cdot\|$ denotes the Euclidean $\ell_2$ norm of a vector. As demonstrated, such distances render CD highly sensitive to outliers  

    \item \underline{\em Hyperbolic distance:} Hyperbolic space is characterized as a homogeneous space with constant negative curvature, where the distance between any two points depends on the curvature passing through them. Typically, there are five isometric models to construct a hyperbolic space, with the Poincaré model being particularly prominent in deep learning applications \cite{9658224}. For instance, if two points $x_{ij}$ and $y_{ik}$ are located within the Poincaré unit ball, meaning $\|x_{ij}\|<1$ and $\|y_{ik}\|<1$, their hyperbolic distance is defined as 
    \begin{align}\label{eqn:hyperbolic}
        d(x_{ij}, y_{ik}) = \arccosh\left(1+2\frac{\|x_{ij}-y_{ik}\|^2}{(1-\|x_{ij}\|^2)(1-\|y_{ik}\|^2)}\right).
    \end{align}
    Note that the hyperbolic distance can be always defined based on $\arccosh$, no matter what model is used to represent the hyperbolic space.
\end{itemize}

\bfsection{Learning Objective for Point Cloud Completion.} Based on the definition of CD in Eq. \ref{eqn:CD}, a simple learning objective can be written as follows:
\begin{align}\label{eqn:obj}
    \min_{\omega\in\Omega}\sum_i F_i(\omega) \stackrel{def}{=} \min_{\omega\in\Omega}\sum_i D(f(x_i;\omega), y_i),
\end{align}
where $\Omega$ denotes the feasible solution space for $\omega$ defined by some constraints such as regularization.

\subsection{Hyperbolic Chamfer Distance}
\bfsection{Challenges.} Listed below are multiple challenges that hinder the direct substitution of Eq. \ref{eqn:hyperbolic} into Eq. \ref{eqn:CD}:
\begin{itemize}[nosep, leftmargin=*]
    \item \underline{\em Domain constraint:} It is essential that the norm of each 3D point remains strictly smaller than 1. To address this, techniques like clipping \cite{zhang2019gradient} can be utilized to help mitigate the issue.

    \item \underline{\em Computational burden:} The calculation in Eq. \ref{eqn:hyperbolic} is considerably more complex than the Euclidean distance, resulting in a significantly higher computational burden, particularly in large-scale settings. Specifically, the matching complexity for each point cloud pair is $O(|x_i||y_i|)$, which is problematic when the number of points in point clouds exceeds 10K. To address this challenge, the hyperbolic distance is often computed in Gyrovector space~\cite{ungar2001hyperbolic,ungar2008analytic,ungar2008gyrovector,yue2023hyperbolic}, a generalization of Euclidean vector spaces based on M\"{o}bius transformations~\cite{geoopt2020kochurov}. Despite this approach, such operations still involve excessive computation, making them inefficient for large-scale settings.
\end{itemize}
We have observed that (1) computing Euclidean distances is significantly faster than calculating hyperbolic distances, and (2) hyperbolic distances are based on $\arccosh$. The question then arises: how can we define the hyperbolic Chamfer distance and optimize its computation?

\subsubsection{Definition}
Inspired by the hyperbolic distance defined in Eq.~\ref{eqn:hyperbolic}, we introduce a novel distance metric, the {\em Hyperbolic Chamfer Distance (\textbf{HyperCD})}, based on Eq. \ref{eqn:CD}. This is defined as 
\begin{align}\label{eqn:hypercd}
    d(x_{ij},y_{ik}) = \arccosh\left(1+\alpha\|x_{ij}-y_{ik}\|^2\right), \alpha>0.
\end{align}
It is important to note that the hyperbolic distance in Eq.~\ref{eqn:hyperbolic} can be viewed as a special case of Eq.~\ref{eqn:hypercd}, where $\alpha = \frac{2}{(1-\|x_{ij}\|^2)(1-\|y_{ik}\|^2)}$ is a function of $x_{ij}$ and $y_{ik}$. Additionally, we present an efficient algorithm for computing the HyperCD between point clouds in Alg. \ref{alg:hypercd}, which exhibits a computational complexity comparable to that of the Euclidean Chamfer Distance.

\subsubsection{Learning with HyperCD as Loss Function}
The learning process in backpropagation for updating network weights can be discussed as follows. Referring to the learning objective in Eq. \ref{eqn:obj}, we obtain the expression:
\begin{align}
    \frac{\partial F_i}{\omega} = \frac{1}{|\Tilde{x}_i|}\sum_j\frac{\partial d(\Tilde{x}_{ij}, y_{im(j)})}{\partial \omega} + \frac{1}{|y_i|}\sum_k\frac{\partial d(\Tilde{x}_{in(k)}, y_{ik})}{\partial \omega}
\end{align}
Here, $\Tilde{x}_i$ represents the output point cloud from the network $f$ when $x_i$ is used as input, while $m(j)$ and $n(k)$ refer to the indices of the nearest neighbor matches.Additionally, the following equation can be expressed as

\begin{align}\label{eqn:derivative}
    \frac{\partial d(\Tilde{x}_{ij}, y_{im(j)})}{\partial \omega} = z_{ij} \cdot \frac{\partial \|\Tilde{x}_{ij} - y_{im(j)}\|}{\partial \omega}
\end{align}

where $z_{ij} = \frac{2\alpha \|\Tilde{x}_{ij} - y_{im(j)}\|}{\sqrt{\left(1+\alpha \|\Tilde{x}_{ij} - y_{im(j)}\|^2\right)^2-1}}\in\mathbb{R}$ represents the weight for the gradient feature in backpropagation. This {\em implicit} weighting mechanism is solely dependent on the Euclidean distances, making it position-aware. The same gradient feature, $\frac{\partial \|\Tilde{x}_{ij} - y_{im(j)}\|}{\partial \omega}$, is employed in both CD and DCD. The key distinction among these distance metrics during learning lies in the weighting mechanism.

\begin{algorithm}[t]
    \SetAlgoLined
    \SetKwInOut{Input}{Input}\SetKwInOut{Output}{Output}
    \Input{a point cloud pair $(x_i,y_i)$, hyperparameter $\alpha>0$}
    \Output{HyperCD $D(x_i,y_i)$}
    \BlankLine
    Initialize a matrix $M$, $D_1\leftarrow0, D_2\leftarrow0$;
    
    \ForEach{$j,k$}
    {
        $M_{jk}\leftarrow\|x_{ij} - y_{ik}\|^2$;
    }
    
    \ForEach{$j$}        
    {
        $D_1\leftarrow D_1 + \arccosh(1+\alpha\min_k M_{jk})$;
    }

    \ForEach{$k$}        
    {
        $D_2\leftarrow D_2 + \arccosh(1+\alpha\min_j M_{jk})$;
    }
    
    \Return $D(x_i,y_i) \leftarrow \frac{D_1}{|x_i|} + \frac{D_2}{|y_i|}$;
    \caption{HyperCD}\label{alg:hypercd}
\end{algorithm}

\subsection{Analysis on HyperCD}
\begin{prop}\label{prop:1}
    Consider $d(x_{ij},y_{ik}) = g(\|x_{ij}-y_{ik}\|)$ in Eq. \ref{eqn:CD} where function $g$ is {\em strictly} increasing. It holds that 
    \begin{align}
        \min_kd(x_{ij},y_{ik}) = g\left(\min_k\|x_{ij}-y_{ik}\|\right).
    \end{align}
\end{prop}
The proposition asserts that for a strictly increasing function $g$, $g$ and $\min$ can be interchanged. As a result, the computational complexity of $g\left(\min_k\|x_{ij}-y_{ik}\|\right)$ is only marginally greater than that of calculating Euclidean distances, with the additional step of computing $g$.

\begin{prop}\label{prop:2}
Let $h(x) = \arccosh(1 + \alpha x^{\beta})$ for all $x \geq 0$. The function $h$ is strictly increasing if and only if $\alpha > 0$ and $\beta > 0$.
\end{prop}
The validity of this proposition follows directly from the fact that the derivative of $h$ is positive when $\alpha > 0$ and $\beta > 0$.

\begin{prop}\label{prop:3}
Considering function $h$ in Prop. \ref{prop:2}, then its derivative, $\frac{\partial h}{\partial x} = \frac{\alpha\beta x^{\beta-1}}{\sqrt{\left(1+\alpha x^{\beta}\right)^2-1}}$, satisfies that
    \begin{align}
        & \lim_{x\rightarrow0^+}\frac{\partial h(x)}{\partial x} = \lim_{x\rightarrow0^+} \frac{\alpha\beta}{\sqrt{2\alpha}}x^{\frac{\beta}{2}-1} = \left\{
            \begin{array}{ll}
                +\infty, & 0<\beta<2 \\
                \frac{\alpha\beta}{\sqrt{2\alpha}}, & \beta = 2 \\
                0, & \beta>2
            \end{array}
        \right. \\
        & \lim_{x\rightarrow+\infty}\frac{\partial h(x)}{\partial x} = \frac{\beta}{x}.
    \end{align}
\end{prop}
As $x$ approaches 0, the function $h$ behaves similarly to a power function, with the exception of the special case where $\beta = 2$, which results in a constant value. This observation aligns with the analysis presented in Fig. \ref{fig:function}~(right), suggesting that only $\beta = 2$ offers an effective weighting mechanism for maintaining accurate point matches. Similarly, as $x$ approaches infinity, all curves corresponding to the same $\beta$ will converge to a power function.

\begin{figure}[t]
    \begin{center}
    \vspace{-3mm}
	\centerline{\includegraphics[width=\linewidth, keepaspectratio,]{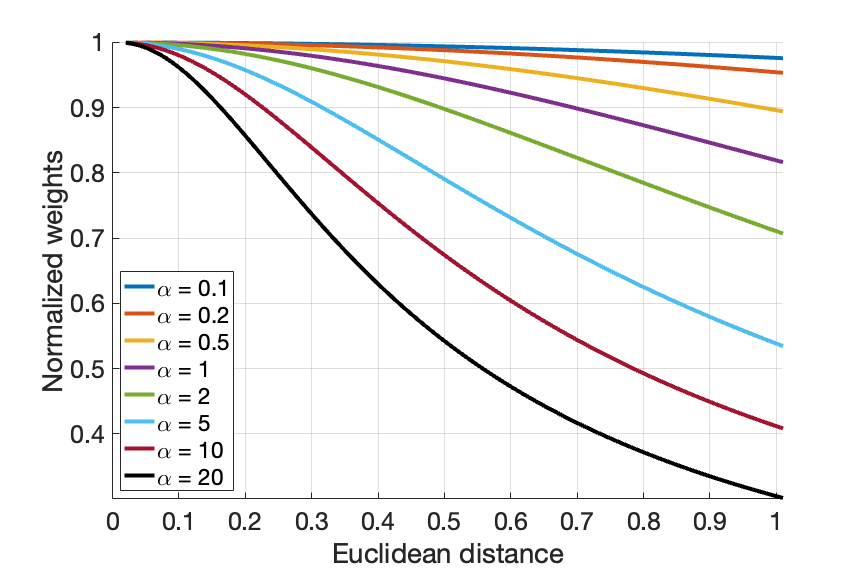}}
	\end{center}
	\vspace{-5mm}
    \caption{Illustration of the gradient weights using our HyperCD. All the numbers are normalized by $\frac{1}{\sqrt{2\alpha}}$.} 
\label{fig:z}
\end{figure}

\begin{prop}
    The weight $z_{ij}$ in Eq. \ref{eqn:derivative} is strictly decreasing \wrt $\|\Tilde{x}_{ij} - y_{im(j)}\|$ for an arbitrary $\alpha>0$.
\end{prop}
Figure \ref{fig:z} demonstrates how the weights change with respect to the distances for various values of $\alpha$. For small values of $\alpha$, such as $\alpha \leq 2$, the curves show a gradual decrease, which may help in preserving good matches more effectively during backpropagation.

\begin{table*}[t]
	\centering	
	\small
	\setlength{\tabcolsep}{4pt}
	\caption{Completion results on PCN in terms of per-point L1 Chamfer distance $\times 1000$ (lower is better).}
	\begin{tabular}{c|c|cccccccc}
		\toprule
		Methods & Average & Plane & Cabinet & Car & Chair & Lamp & Couch & Table & Boat  \\
		\midrule
		FoldingNet \cite{yang2018foldingnet} & 14.31 & 9.49 & 15.80 & 12.61 & 15.55 & 16.41 & 15.97 & 13.65 & 14.99 \\
		TopNet \cite{topnet} & 12.15 & 7.61 & 13.31 & 10.90 & 13.82 & 14.44 & 14.78 & 11.22 & 11.12 \\
		AtlasNet \cite{groueix2018papier} & 10.85 & 6.37 & 11.94 & 10.10 & 12.06 & 12.37 & 12.99 & 10.33 & 10.61 \\
		GRNet \cite{GRNet} & 8.83 & 6.45 & 10.37 & 9.45 & 9.41 & 7.96 & 10.51 & 8.44 & 8.04 \\
		CRN \cite{wang2020cascaded} & 8.51 & 4.79 & 9.97 & 8.31 & 9.49 & 8.94 & 10.69 & 7.81 & 8.05 \\
		NSFA \cite{zhang2020detail} & 8.06 & 4.76 & 10.18 & 8.63 & 8.53 & 7.03 & 10.53 & 7.35 & 7.48 \\
		FBNet \cite{yan2022fbnet} & 6.94 & 3.99 & 9.05 & 7.90 & 7.38 & 5.82 & 8.85 & 6.35 & 6.18 \\

        \midrule 
		PCN \cite{pcn} & 11.27 & \textbf{5.50} & 22.70 & 10.63 & \textbf{8.70} & \textbf{11.00} & \textbf{11.34} & 11.68 & \textbf{8.59} \\

		{\bf HyperCD + PCN} & \textbf{10.59} & 5.95 & \textbf{11.62} & \textbf{9.33} & 12.45 & 12.58 & 13.10 & \textbf{9.82} & 9.85 \\
        
        \midrule 
		FoldingNet \cite{yang2018foldingnet} & 14.31 & 9.49 & 15.80 & 12.61 & 15.55 & 16.41 & 15.97 & 13.65 & 14.99 \\

        {\bf HyperCD + FoldingNet} & \textbf{12.09} & \textbf{7.89} & \textbf{12.90} & \textbf{10.67} & \textbf{14.55} & \textbf{13.87} & \textbf{14.09} & \textbf{11.86} & \textbf{10.89} \\

        \midrule 
		PMP-Net \cite{wen2021pmp} & 8.73 & 5.65 & 11.24 & 9.64 & 9.51 & 6.95 & 10.83 & 8.72 & 7.25 \\

        {\bf HyperCD + PMP-Net} & \textbf{8.40} & \textbf{5.06} & \textbf{10.67} & \textbf{9.30} & \textbf{9.11} & \textbf{6.83} & \textbf{11.01} & \textbf{8.18} & \textbf{7.03} \\

        \midrule         
		PoinTr \cite{pointtr} & 8.38 & 4.75 & 10.47 & 8.68 & 9.39 & 7.75 & 10.93 & 7.78 & 7.29 \\
        {\bf HyperCD + PoinTr} & \textbf{7.56} & \textbf{4.42} & \textbf{9.77} & \textbf{8.22} & \textbf{8.22} & \textbf{6.62} & \textbf{9.62} & \textbf{6.97} & \textbf{6.67} \\
        
        \midrule 

		SnowflakeNet \cite{snowflakenet} & 7.21 & 4.29 & 9.16 & 8.08 & 7.89 & 6.07 & 9.23 & 6.55 & 6.40 \\

        {\bf HyperCD + SnowflakeNet} & \textbf{6.91} & \textbf{3.95} & \textbf{9.01} & \textbf{7.88} & \textbf{7.37} & \textbf{5.75} & \textbf{8.94} & \textbf{6.19} & \textbf{6.17} \\

        \midrule 
		PointAttN \cite{wang2022pointattn} & 6.86 & 3.87 & 9.00 & 7.63 & 7.43 & 5.90 & 8.68 & 6.32 & 6.09  \\
		DCD + PointAttN & 7.54 & 4.47 & 9.65 & 8.14 & 8.12 & 6.75 & 9.60 & 6.92 & 6.67   \\

        {\bf HyperCD + PointAttN} & \textbf{6.68} & \textbf{3.76} & \textbf{8.93} & \textbf{7.49} & \textbf{7.06} & \textbf{5.61} & \textbf{8.48} & \textbf{6.25} & \textbf{5.92} \\
  
		\midrule

        SeedFormer \cite{zhou2022seedformer} & 6.74 & 3.85 & 9.05 & 8.06 & 7.06 & 5.21 & 8.85 & 6.05 & 5.85 \\

        {DCD + SeedFormer} & 24.52 & 16.42 & 26.23 & 21.08 & 20.06& 18.30 & 26.51 & 18.23 & 18.22 \\       
        {\bf HyperCD + SeedFormer} & \textbf{6.54} & \textbf{3.72} & \textbf{8.71} & \textbf{7.79} & \textbf{6.83} & \textbf{5.11} & \textbf{8.61} & \textbf{5.82} & \textbf{5.76} \\

  \bottomrule
	\end{tabular}

	\label{table:pcn}
\end{table*}

\begin{figure*}[t]
	\begin{center}
		\includegraphics[width=.97\linewidth]{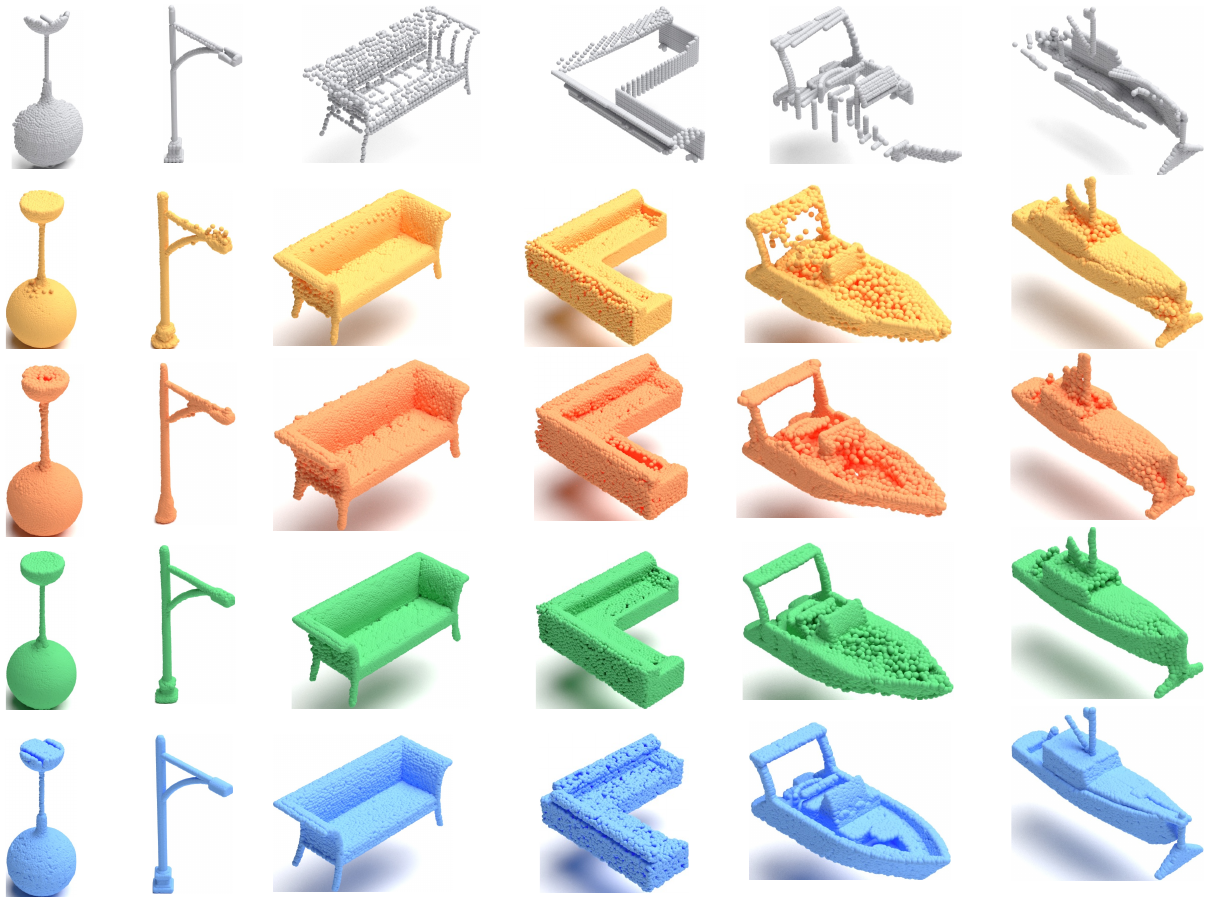}
	\end{center}
	\caption{Visual comparison of point cloud completion results on PCN. {\bf Row-1:} Inputs of incomplete point clouds. {\bf Row-2:} Outputs of PointAttN with CD. {\bf Row-3:} Outputs of PointAttN with DCD. {\bf Row-4:} Outputs of PointAttN with HyperCD. {\bf Row-5:} Ground truth.}
	\label{fig:pcn}
\end{figure*}

\begin{table*}[t]
	\centering	
	\small
	\caption{Completion results on ShapeNet-55 based on L2 Chamfer distance $\times 1000$ (lower is better) and F-Score@1$\%$ (higher is better).}
	\begin{tabular}{c|ccccc|cccc|c}
		\toprule
		Methods & Table & Chair & Plane & Car & Sofa & CD-S & CD-M & CD-H & CD-Avg & F1  \\
		\midrule
        PFNet \cite{PFNet}  				& 3.95 & 4.24 & 1.81 & 2.53 & 3.34 & 3.83 & 3.87 & 7.97 & 5.22 & 0.339 \\ 
		FoldingNet \cite{yang2018foldingnet}  	& 2.53 & 2.81 & 1.43 & 1.98 & 2.48 & 2.67 & 2.66 & 4.05 & 3.12 & 0.082 \\
        TopNet \cite{topnet}  		& 2.21 & 2.53 & 1.14 & 2.18 & 2.36 & 2.26 & 2.16 & 4.3 & 2.91 & 0.126  \\
        PCN \cite{pcn}  				& 2.13 & 2.29 & 1.02 & 1.85 & 2.06 & 1.94 & 1.96 & 4.08 & 2.66 & 0.133 \\
        GRNet \cite{GRNet}  				& 1.63 & 1.88 & 1.02 & 1.64 & 1.72 & 1.35 & 1.71 & 2.85 & 1.97 & 0.238 \\
        PoinTr \cite{pointtr}  			& 0.81 & 0.95 & 0.44 & 0.91 & 0.79 & 0.58 & 0.88 & 1.79 & 1.09 & 0.464 \\
        \midrule
        SeedFormer \cite{zhou2022seedformer} & 0.72 & 0.81 & 0.40 & 0.89 & 0.71 & 0.50 & 0.77 & 1.49 & 0.92 & 0.472 \\        		
		{\bf HyperCD + SeedFormer} & \textbf{0.66} & \textbf{0.74} & \textbf{0.35} & \textbf{0.83} & \textbf{0.64} & \textbf{0.47} & \textbf{0.72} & \textbf{1.40} & \textbf{0.86} & \textbf{0.482} \\
		\bottomrule
	\end{tabular}

	\label{table:shapenet55}
\end{table*}

\begin{table*}[t]
	\centering
	\small
	\caption{Completion results on ShapeNet-34 based on L2 Chamfer distance $\times 1000$ (lower is better) and F-Score@1$\%$ (higher is better).}
	\begin{tabular}{c|ccccc|ccccc}
		\toprule
		\multirow{2}{*}{Methods} & \multicolumn{5}{c|}{34 seen categories} & \multicolumn{5}{c}{21 unseen categories} \\
		        & CD-S & CD-M & CD-H & CD-Avg & F1 & CD-S & CD-M & CD-H & CD-Avg & F1 \\
		\midrule
        PFNet \cite{PFNet}  				 & 3.16 & 3.19 & 7.71 & 4.68 & 0.347 & 5.29 & 5.87 & 13.33 & 8.16 & 0.322 \\
		FoldingNet \cite{yang2018foldingnet}  	 & 1.86 & 1.81 & 3.38 & 2.35 & 0.139 & 2.76 & 2.74 & 5.36 & 3.62 & 0.095 \\
        TopNet \cite{topnet}  		 & 1.77 & 1.61 & 3.54 & 2.31 & 0.171 & 2.62 & 2.43 & 5.44 & 3.50 & 0.121 \\ 
        PCN \cite{pcn}  				 & 1.87 & 1.81 & 2.97 & 2.22 & 0.154 & 3.17 & 3.08 & 5.29 & 3.85 & 0.101 \\
        GRNet \cite{GRNet}  				 & 1.26 & 1.39 & 2.57 & 1.74 & 0.251 & 1.85 & 2.25 & 4.87 & 2.99 & 0.216 \\
        PoinTr \cite{pointtr}   			 & 0.76 & 1.05 & 1.88 & 1.23 & 0.421 & 1.04 & 1.67 & 3.44 & 2.05 & 0.384 \\
        \midrule[0.3pt]
        SeedFormer \cite{zhou2022seedformer}    & 0.48 & 0.70 & 1.30 & 0.83 & 0.452 & 0.61 & 1.08 & 2.37 & 1.35 & 0.402 \\        		
		{\bf HyperCD + SeedFormer} & \textbf{0.46} & \textbf{0.67} & \textbf{1.24} & \textbf{0.79} & \textbf{0.459} & \textbf{0.58} & \textbf{1.03} & \textbf{2.24} & \textbf{1.31} & \textbf{0.428} \\
		\bottomrule
	\end{tabular}

	\label{table:shapenet34}
\end{table*}

\section{Experiments}
\label{sec:experiment}
\bfsection{Datasets.}
We conduct verification and analysis of our HyperCD for tasks such as point cloud completion, point cloud self-supervision pre-training, single image reconstruction, and upsampling, using the following representative benchmark datasets. 
\begin{itemize}[nosep, leftmargin=*]
    \item {\em ShapeNet-Part:} The ShapeNet-Part benchmark \cite{yi2016scalable}, a subset of ShapeNetCore \cite{chang2015shapenet} 3D meshes, includes 17,775 distinct 3D meshes distributed across 16 categories. Ground truth point cloud data is generated by uniformly sampling 2,048 points on each mesh. Partial point cloud data is created by selecting a viewpoint at random from multiple options, designating it as the center, and removing points within a specified radius from the complete dataset. This process is similar to the creation of the ShapeNet-55/34 \cite{pointtr} benchmark in PoinTr. A total of 512 points are removed from each point cloud.

    \item {\em PCN:} The PCN dataset \cite{pcn} is widely regarded as one of the most popular benchmark datasets for point cloud completion. It is a subset of ShapeNet \cite{chang2015shapenet}, encompassing shapes from eight distinct categories. Incomplete point clouds are generated by back-projecting 2.5D depth images captured from eight viewpoints, thereby simulating real-world sensor data. For each shape, 16,384 points are uniformly sampled from the mesh surfaces to represent the complete ground truth, while 2,048 points are sampled as partial input \cite{pcn,zhou2022seedformer}.

    \item {\em ShapeNet-55/34:} The ShapeNet-55 and ShapeNet-34 datasets are derived from the synthetic ShapeNet \cite{chang2015shapenet} dataset, but they encompass a broader range of object categories and feature incomplete patterns. ShapeNet-55 includes all 55 categories from ShapeNet, with a total of 41,952 shapes for training and 10,518 shapes for testing. In contrast, ShapeNet-34 focuses on a subset of 34 categories for training, leaving 21 categories unseen for testing. Specifically, ShapeNet-34 consists of 46,765 object shapes for training, 3,400 shapes for testing on seen categories, and 2,305 shapes for testing on novel (unseen) categories. Both datasets sample 2,048 points as input and 8,192 points as ground truth. Consistent with the evaluation strategy outlined in \cite{pointtr}, eight fixed viewpoints are selected, and the number of points in the partial point cloud is set to 2,048, 4,096, or 6,144, corresponding to three difficulty levels—\emph{simple}, \emph{moderate}, and \emph{hard}—during the test phase.

     \item {\em KITTI:} The KITTI dataset \cite{geiger2013vision} consists of a series of real-world Velodyne LiDAR scans, also sourced from the PCN dataset \cite{pcn}. For each frame, car objects are identified based on 3D bounding boxes, leading to a total of 2,401 partial point clouds. These point clouds in KITTI are sparse, lacking complete point clouds as ground truth.

     \item {\em SVR ShapeNet:} The SVR ShapeNet\cite{xiang2022snowflake}, derived from the synthetic ShapeNet\cite{chang2015shapenet}, includes 43,783 shapes across 13 categories. It is divided into training, testing, and validation sets, with respective proportions of 70\%, 20\%, and 10\%.

     \item {\em PU1K:} PU1K \cite{qian2021pu} significantly surpasses the previous largest publicly available dataset from PU-GAN \cite{li2019pu}, being nearly eight times its size. Comprising 1,147 3D models, PU1K is divided into 1,020 training samples and 127 testing samples. The training set includes 120 models sourced from PU-GAN's dataset \cite{li2019pu}, supplemented by 900 distinct models from ShapeNetCore \cite{chang2015shapenet}. For the testing set, 27 models come from PU-GAN, while the remaining 100 models are sourced from ShapeNetCore. These ShapeNetCore models are selected from 50 categories, with 200 models chosen at random from each, totaling 1,000 unique models. These models vary in shape complexity, ensuring a broad range of diversity. PU1K thus encompasses a wide semantic spectrum of 3D objects, including both simple and intricate shapes.



\end{itemize}

\bfsection{Implementation.}
For the point cloud completion task, we select three state-of-the-art networks—CP-Net \cite{lin2022cosmos}, PointAttN \cite{wang2022pointattn}, and SeedFormer \cite{zhou2022seedformer}—as our backbone models for comparison and analysis. Additionally, we include results from several well-established baselines to assess the generality of HyperCD. All networks, both with and without HyperCD, are trained from scratch using PyTorch \cite{paszke2019pytorch}. Each of these models is designed with a supervision signal based on the dissimilarity between generated and ground truth point clouds. In the case of the lightweight CP-Net, which employs a single stage of supervision using dissimilarity between prediction and ground truth, we replace its original loss function with ours, as well as others, to validate our approach and analyze the performance differences across the various loss functions.
In the case of multi-stage architectures, such as PointAttN and SeedFormer, loss functions are incorporated at each stage, advancing from coarse to fine point clouds. HyperCD replaces the loss functions in all stages, allowing it to be involved throughout the training process. To ensure a fair comparison, all other loss functions evaluated alongside HyperCD in this paper are handled consistently with HyperCD.


For the single-view reconstruction task, we substitute CD with HyperCD in SnowflakeNet\cite{snowflakenet} for training while maintaining the original architecture to ensure a fair comparison.

For the point cloud upsampling task, we train PU-Net\cite{yu2018pu}, 3PU\cite{yifan2019patch}, PU-GCN\cite{li2019pu}, and PUCRN\cite{du2022point} using HyperCD. Consistent with other tasks, we maintain the original architecture to ensure a fair comparison.



The consistency of all hyperparameters, such as learning rate, batch size, and training epochs, is preserved in alignment with the baseline settings to ensure a fair comparison. The tuning engineer's role is focused on the grid search for the hyperparameter $\alpha$ in HyperCD. Given the extensive number of methods involved, the experiments are conducted across multiple servers: one with 10 NVIDIA RTX 2080Ti 11G GPUs, another with 4 NVIDIA A100 80G GPUs, and a third with 4 NVIDIA V100 16G GPUs.


\bfsection{Evaluation Metrics.}
To ensure a fair comparison in completion tasks, we assess the performance of all methods using CD. Additionally, F1-Score@1\% \cite{tatarchenko2019single} is utilized to evaluate ShapeNet-55/34 under the same experimental settings as in the literature. For a more comprehensive comparison, we also present the original results of several other methods on PCN and ShapeNet-55/34. CD is employed for SVR, as the reconstruction results are also point clouds. For point cloud unsampling, we adopt CD, Hausdorff distance (HD), and point-to-surface distance (P2F) with respect to ground truth meshes, in line with previous work, as evaluation metrics.

\subsection{State-of-the-art Comparison}
\label{sec:Main_exp}
\bfsection{PCN.} 
As outlined in the literature, we present the CD with L1-distance results in Table \ref{table:pcn}, broken down by category. Additionally, we include the results from training with DCD. Notably, replacing the HyperCD loss allows two of the baselines to surpass their previous state-of-the-art performance, although there is a slight decrease in performance when DCD is utilized. We also report several baseline results with the HyperCD loss replacement, all of which show consistent improvements. As previously discussed, numerical metrics (e.g., CD) may not always accurately reflect visual quality; therefore, we provide qualitative evaluation results in Fig.~\ref{fig:pcn}, comparing them to those generated by the baseline models using CD and DCD loss functions. Both models generally reconstruct point clouds with a similar overall shape, but the results using CD tend to exhibit distortions in areas with high noise on the surface. With the integration of HyperCD loss during training, the baseline network produces a more accurate reconstruction of the point cloud's overall shape, while preserving realistic details and significantly reducing surface noise.
While DCD demonstrates superior capability in controlling the noise level of generated point clouds, it struggles to maintain fine details and is also prone to distortion. 


\bfsection{ShapeNet-55/34.} 
We also conduct testing on the ShapeNet-55 dataset, which consists of 55 object categories across three levels of incompleteness (difficulty), to assess HyperCD's adaptability to tasks with greater diversity. Table~\ref{table:shapenet55} presents the average L2 Chamfer distances for each of the three difficulty levels, as well as the overall CDs. Consistent with prior conventions, results for five categories (Table, Chair, Plane, Car, and Sofa), which contain over 2,500 samples in the training set, are provided. Complete results for all 55 categories can be found in the supplemental material. Additionally, results under the F-Score@1$\%$ metric are included. As shown in Table~\ref{table:shapenet55}, incorporating HyperCD leads to a noticeable improvement in baseline performance. For a more intuitive evaluation of the reconstructed results, we also include qualitative assessments in the supplemental materials, comparing them to the baseline results.
While the reconstructed results demonstrate superior numerical performance, the model trained with HyperCD excels in surface area reconstruction and detail preservation with reduced noise. The enhancement in both numerical and qualitative evaluations suggests that HyperCD is well-equipped to handle point completion tasks with high diversity.

In the evaluation on ShapeNet-34, we assess the performance across 34 seen categories (those used in training) and 21 unseen categories (not included in training). As shown in Table~\ref{table:shapenet34}, HyperCD demonstrates its ability to enhance the baseline model's performance by achieving higher scores. This improvement highlights the high generalizability of our loss function for point cloud completion tasks, effectively handling both seen and unseen categories.

\bfsection{KITTI.} 
Our evaluation of HyperCD's efficacy in real-world scenarios involves fine-tuning two baseline models, utilizing ShapeNetCars as a training dataset, following the protocol established by GRNet, and subsequently testing their performance on the KITTI dataset. The resulting metrics, including Fidelity and MMD, are presented in Table~\ref{tab:KITTI}. Notably, HyperCD consistently enhances the performance of these baseline models.

\begin{table*}
\caption{Results on LiDAR scans from KITTI dataset under the  Fidelity and MMD metrics.} \small
\label{tab:KITTI}
\centering
\begin{tabular}[\linewidth]{l | c c| c c |c c }
\toprule
CD-$\ell_2$ ($\times$ 1000) & FoldingNet & HyperCD + F. & PoinTr &  HyperCD + P. & \\
\midrule
Fidelity $\downarrow$ & 7.467 &  \textbf{2.214} & 0.000  & \textbf{0.000} \\
MMD $\downarrow$ & 0.537 & \textbf{0.386} & 0.526& \textbf{0.507} \\
\bottomrule
\end{tabular}
\end{table*}

            

\bfsection{SVR.} 
We also includes an alternative task utilizing HyperCD, specifically {\em single view reconstruction (SVR)}, which seeks to reconstruct a point cloud from a captured image of the underlying object. In this context, we adopt a similar approach to that employed by 3DAttriFlow and SnowflakeNet, wherein we sample 30k points from the watertight mesh in ShapeNet as ground truth, and subsequently output 2048 points for evaluation based on per-point L1-CD$\times10^2$. Notably, we replace CD in SnowflakeNet with HyperCD for training purposes, and provide average comparison results in Table \ref{table:svr}, which demonstrate the effectiveness of HyperCD across different tasks.

\bfsection{Upsampling.} 
Table \ref{tab:sota_comparasion_PU1K} provides a comprehensive comparison of state-of-the-art methods, including PU-Net \cite{yu2018pu}, 3PU \cite{yifan2019patch}, PU-GCN \cite{li2019pu}, and PUCRN \cite{du2022point}, on the PU1K dataset. Notably, we do not include PU-GAN in this evaluation due to the reasons outlined in PU-GCN \cite{qian2021pu}. The results demonstrate that all considered methods exhibit lower performance overall when trained and evaluated using HyperCD, particularly in CD and HD metrics. Consequently, the PU1K dataset presents a significant challenge to state-of-the-art methods, which contrasts with the relatively easier conditions of the smaller PU-GAN's dataset. Furthermore, PU-GCN outperforms PU-Net and 3PU in all three metrics on this novel challenging dataset.


\begin{table}[t]
\caption{\textbf{Comparison of PU-GCN vs. state-of-the-art on PU1K.} 
PU-GCN exhibits superior performance compared to both PU-Net and 3PU, denoted as bold in the results.
}
\label{tab:sota_comparasion_PU1K}
\centering
\resizebox{0.9\columnwidth}{!}{%
\begin{tabular}{l|ccc}
\toprule
\multirow{2}{*}{\textbf{Network}}
& \multicolumn{1}{c}{\textbf{CD}$\downarrow$} & \multicolumn{1}{c}{\textbf{HD}$\downarrow$} & \multicolumn{1}{c}{\textbf{P2F}$\downarrow$} \\
&\multicolumn{1}{c}{$10^{-3}$} & \multicolumn{1}{c}{$10^{-3}$} & \multicolumn{1}{c}{$10^{-3}$} \\
\midrule
PU-Net\cite{yu2018pu} & 1.155          & 15.170         & 4.834          \\

\textbf{HyperCD + PU-Net}   & \textbf{1.095} & \textbf{14.490} & \textbf{3.730}   \\
\midrule
3PU\cite{yifan2019patch}    & 0.935          & 13.327         & 3.551          \\
\textbf{HyperCD + 3PU}   & \textbf{0.902} & \textbf{12.139} & \textbf{3.386}   \\

\midrule
\textbf{PU-GCN\cite{li2019pu}}                        & 0.585 & \textbf{7.577} & 2.499 \\
\textbf{HyperCD + PU-GCN}   & \textbf{0.572} & 7.584 & \textbf{2.376}   \\

\midrule
PUCRN\cite{du2022point}    & 0.471          & 7.123         & 1.925          \\
\textbf{HyperCD + PUCRN}   & \textbf{0.460} & \textbf{6.597} & \textbf{1.690}   \\

\bottomrule
\end{tabular}
}
\vspace{-10pt}
\end{table}

\setlength{\intextsep}{0pt}
\setlength{\columnsep}{0pt}
\begin{table*}[h]
\small
\centering
\setlength{\tabcolsep}{2pt}
\caption{SVR performance on ShapeNet in terms of per-point L1 Chamfer distance $\times 10^2$ (lower is better).}
\begin{tabular}{c|c|ccccccccccccc}
\toprule
Method &Average &Plane &Bench &Cabinet &Car &Chair &Display &Lamp &Loud. &Rifle &Sofa &Table &Tele. &Vessel\\ \midrule
3DR2N2\cite{choy20163d}     &5.41 &4.94 &4.80 &4.25 &4.73 &5.75 &5.85 &10.64 &5.96 &4.02 &4.72 &5.29 &4.37 &5.07\\
PSGN\cite{fan2017point}       &4.07 &2.78 &3.73 &4.12 &3.27 &4.68 &4.74 &5.60  &5.62 &2.53 &4.44 &3.81 &3.81 &3.84\\
Pixel2mesh\cite{wang2018pixel2mesh} &5.27 &5.36 &5.14 &4.85 &4.69 &5.77 &5.28 &6.87  &6.17 &4.21 &5.34 &5.13 &4.22 &5.48\\
AtlasNet\cite{groueix2018papier}   &3.59 &2.60 &3.20 &3.66 &3.07 &4.09 &4.16 &4.98  &4.91 &2.20 &3.80 &3.36 &3.20 &3.40\\
OccNet\cite{mescheder2019occupancy}     &4.15 &3.19 &3.31 &3.54 &3.69 &4.08 &4.84 &7.55  &5.47 &2.97 &3.97 &3.74 &3.16 &4.43\\ 
Pix2Vox \cite{xie2019pix2vox} & 4.28 & 3.48 & 4.47 & 4.39 & 3.56 & 4.04 & 4.47 & 5.66 & 5.10 & 3.80 & 4.37 & 4.29 & 3.84 & 4.14 \\
Pix2Vox++ \cite{xie2020pix2vox++} & 4.17 & 3.65 & 4.40 & 3.99 & 3.48 & 3.97 & 4.40 & 5.63 & 4.84 & 3.78 & 4.12 & 4.01 & 3.68 & 4.28 \\
DPM \cite{luo2021diffusion} & 3.76 & 2.64 & 3.56 & 3.46 & 3.23 & 4.15 & 4.35 & 5.18 & 5.14 & 2.41 & 4.15 & 3.71 & 3.42 & 3.52 \\

3DAttriFlow \cite{wen20223d} & 3.02 & 2.11 & 2.71 & 2.66 & 2.50 & 3.33 & 3.60 & 4.55 & 4.16 & 1.94 & 3.24 & 2.85 & 2.66 &  2.96 \\


\midrule
SnowflakeNet \cite{snowflakenet}       &2.86 &1.99 &2.54 &2.52 &2.44 &3.13 &3.37 &4.34  &3.98 &1.84 &3.09 &2.71 &2.45 &2.80\\


HyperCD+SnowflakeNet       &\textbf{2.73} &\textbf{1.86} &\textbf{2.48} &\textbf{2.50} &\textbf{2.42} &\textbf{3.11} &\textbf{3.35} &\textbf{4.31}  &\textbf{3.96} &\textbf{1.81} &\textbf{3.07} &\textbf{2.62} &\textbf{2.33} &\textbf{2.71}\\


\bottomrule
\end{tabular}
\label{table:svr}
\end{table*}

\subsection{Analysis}

\begin{table}[t]
	\centering
	\small
	\caption{Completion results of CP-Net with different losses on ShapeNet-Part in terms of per-point L2 Chamfer distance $\times 1000$.}
        \begin{tabular}{c|ccc}
    		\toprule
    		Loss function & CD-Avg \\
    		\midrule
    		L1-CD & 4.16 \\
    		L2-CD & 4.82 \\      
    		DCD & 5.74 \\     
            \midrule
            $y=\arccosh(1+x)$ & 4.43 \\        
            $y=\arccosh(1+x^3)$ & 4.22 \\
            Hyperbolic Distance & 4.09 \\ 
            {\bf HyperCD} & {\bf 4.03} \\
    		\bottomrule
	    \end{tabular}

	\label{table:Shapenet-Part_analysis}
\end{table}
\label{sec:Analysis_exp}

ShapeNet-Part was selected as the primary dataset for comparative analysis with various loss functions. This dataset comprises 16 categories of objects, which is deemed sufficient for our specific analytical needs. The chosen model for this study is a lightweight neural network known as CP-Net \cite{lin2022cosmos}.

\begin{figure}[t]
  \centering
  \resizebox{\columnwidth}{!}{\includegraphics{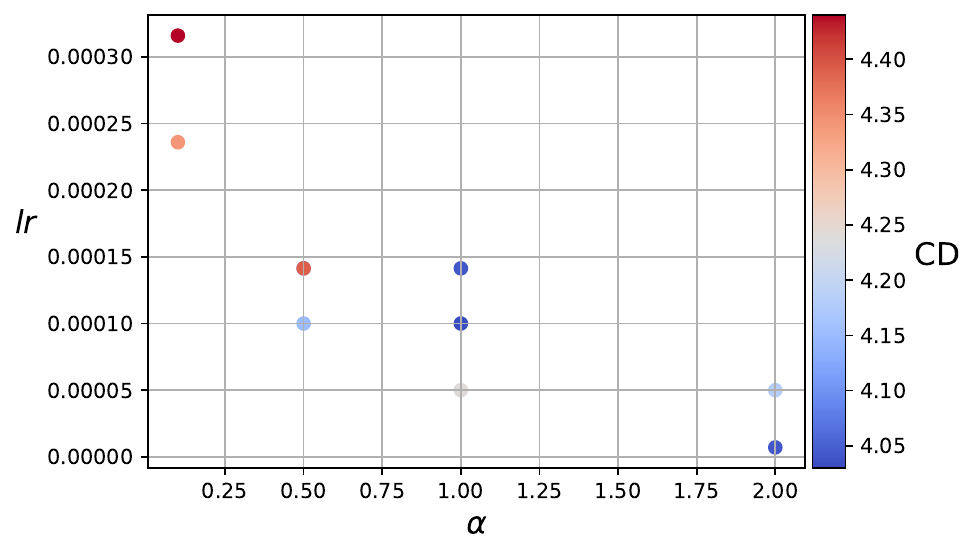}}
  \captionof{figure}{The L1-CD with different $\alpha$ and $lr$.}
  \label{fig:visual_beta_lr}
\end{figure}

\bfsection{Hyperparameters.} 
We demonstrate the correlation between $\alpha$ and learning rate ($lr$) on training performance, as illustrated in Fig.~\ref{fig:visual_beta_lr}. Supplementing our previous discussion regarding $\arccosh$, we conducted experiments to evaluate the efficacy of various $\arccosh$-family metrics (as depicted in Fig.~\ref{fig:function}) on CP-Net, with results summarized in Table~\ref{table:Shapenet-Part_analysis}. Additionally, performance comparisons for several prevalent loss functions are provided in Table~\ref{table:Shapenet-Part_analysis}, highlighting HyperCD's substantial advantage over CD and DCD.

\bfsection{Computation.} Note that our HyperCD has only one additional operation, $\arccosh$, compared to CD and thus in theory both computational efficiency should be comparable betweem the two, but both are much simpler than hyperbolic distance. Numerically, it takes 0.4239$\pm$0.0019, 0.4298$\pm$0.0014 and 0.5335$\pm$0.0368 second per iteration for training CP-Net with CD, HyperCD and hyperbolic distance, respectively.

\begin{figure}[t]
    \begin{center}
	\centerline{\includegraphics[width=\linewidth, keepaspectratio,]{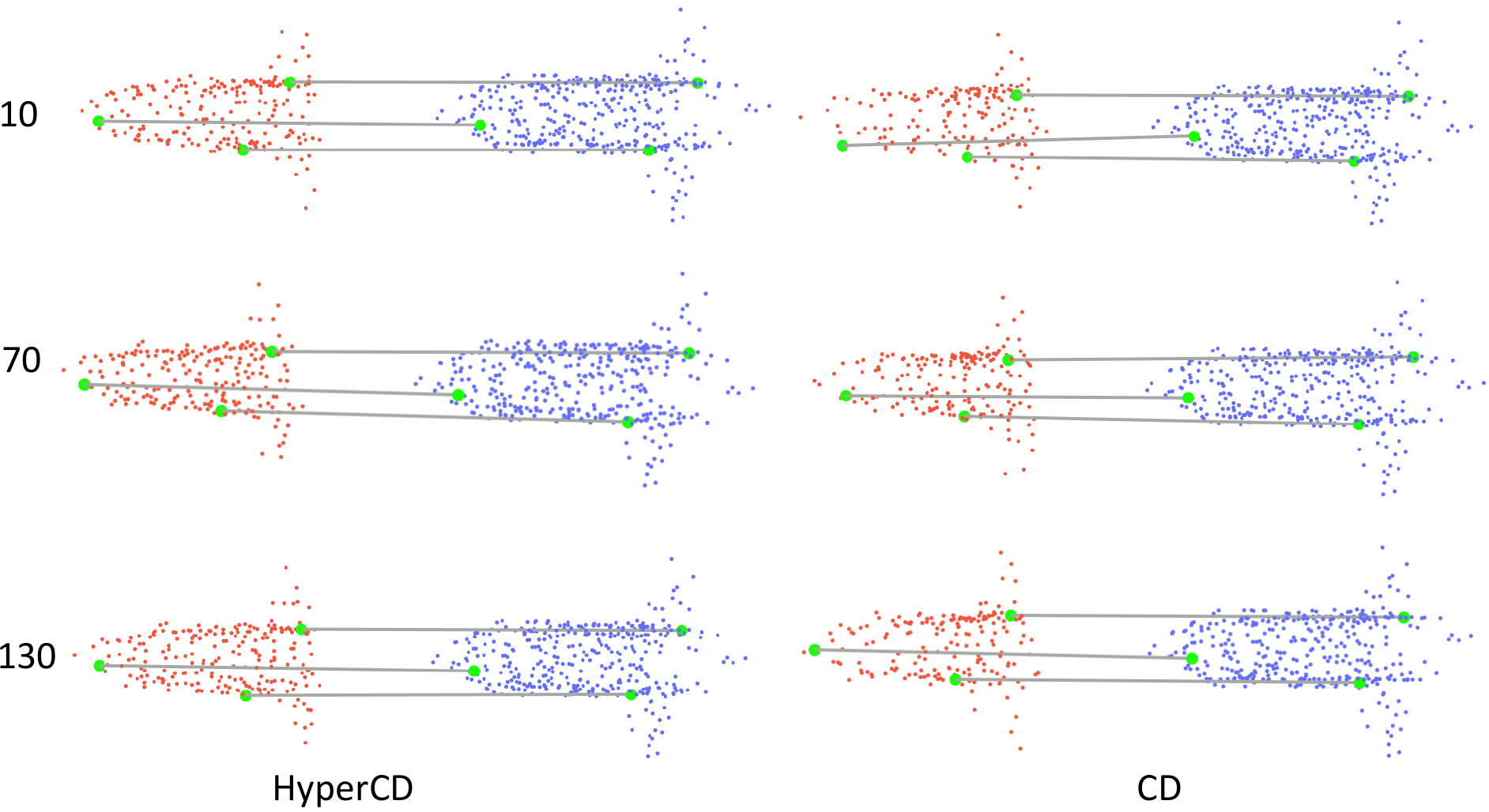}}
	\end{center}
	\vspace{-5.5mm}
    \caption{Illustration of point correspondence change over epochs.} 
\label{fig:point_correspondence}
\end{figure}

\bfsection{Point Correspondences.} 
We also provide an visualization on how the point correspondences change during training. We plot some point correspondences over epochs (10,70,130), as illustrated in Fig. \ref{fig:point_correspondence} where the blue points are ground truth and the red ones are predictions. HyperCD facilitating faster stabilization of correct correspondences and superior convergence compared to CD. Also, as visualized in Fig. \ref{fig:point_correspondence} at the 10-th epoch, the predicted airplane head (red) exhibits notable smoothness and geometric proximity to the ground truth (blue) when employing HyperCD versus CD, underscoring the capacity of HyperCD for preserving geometric features.

\section{Conclusion}


In this work, we investigate the integration of hyperbolic space into point cloud completion and generation tasks, leveraging its exponential growth properties to enhance the Chamfer Distance (CD) loss. Building on insights from hyperbolic distance and recent advancements in CD, a novel distance measure—Hyperbolic Chamfer Distance (HyperCD)—is introduced to address the common outlier issues associated with CD. The effectiveness of HyperCD is validated on multiple benchmark datasets using popular networks, achieving state-of-the-art results in point cloud completion. Additionally, the proposed loss function is adapted for single-view reconstruction and point cloud upsampling, consistently outperforming baseline CD metrics and showcasing its versatility across diverse point cloud tasks. 

\section*{Acknowledgment}
This work was supported by Worcester Polytechnic Institute Internal Fund. Dr. Ziming Zhang were supported partially by NSF CCF-2006738. Dr. Kazunori D Yamada was supported in part by the Top Global University Project from the Ministry of Education, Culture, Sports, Science, and Technology of Japan (MEXT). Computations were partially performed on the NIG supercomputer at ROIS National Institute of Genetics.

\ifCLASSOPTIONcaptionsoff
  \newpage
\fi

\vspace{-1em}

\bibliographystyle{IEEEtran}
\bibliography{egbib}


\vskip -25pt plus -1fil

\begin{IEEEbiography}[{\includegraphics[width=1in,height=1.25in,clip,keepaspectratio]{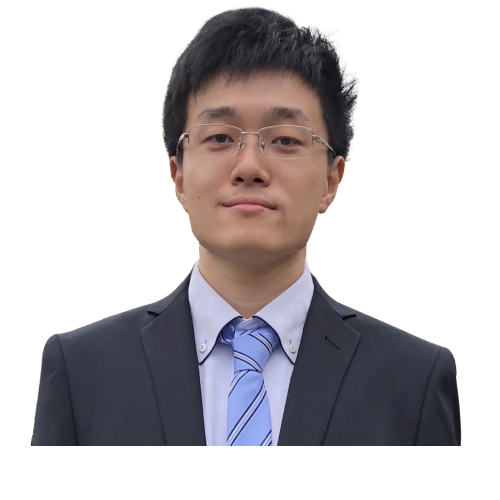}}]
{Fangzhou Lin}
received his M.S. and Ph.D. in information sciences and data science from Tohoku University, Japan, Sendai, in 2021 and 2024.  His research interests include 3D point cloud and medical imaging applications.

\end{IEEEbiography}

\vskip -40pt plus -1fil

\begin{IEEEbiography}[{\includegraphics[width=1in,height=1.25in,clip,keepaspectratio]{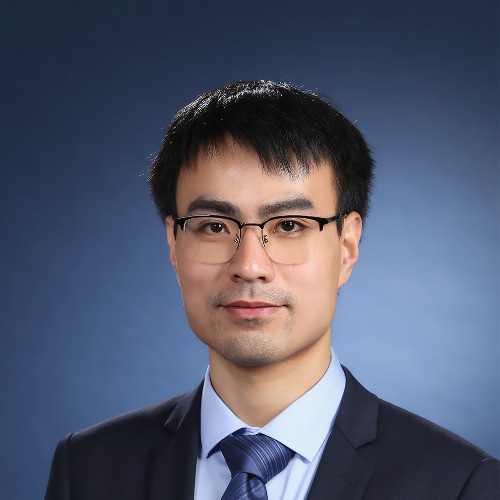}}]{Songlin Hou}
 is a Software Engineer at Dell Technologies and has held research positions at Worcester Polytechnic Institute and The Hong Kong Polytechnic University. He earned his Master of Science degrees in Data Science from The University of Texas at Austin and in Computer Science from Worcester Polytechnic Institute. His research interests include machine learning, deep learning, and computer vision. He has also been recognized with several prestigious awards, including the Finalist Honor in Dell ISG Global Hackathon Competition in 2021 and the 1st Prize in the Intel International Embedded System Design Competition in 2014. 

\end{IEEEbiography}

\vskip -30pt plus -1fil

\begin{IEEEbiography}[{\includegraphics[width=1in,height=1.25in,clip,keepaspectratio]{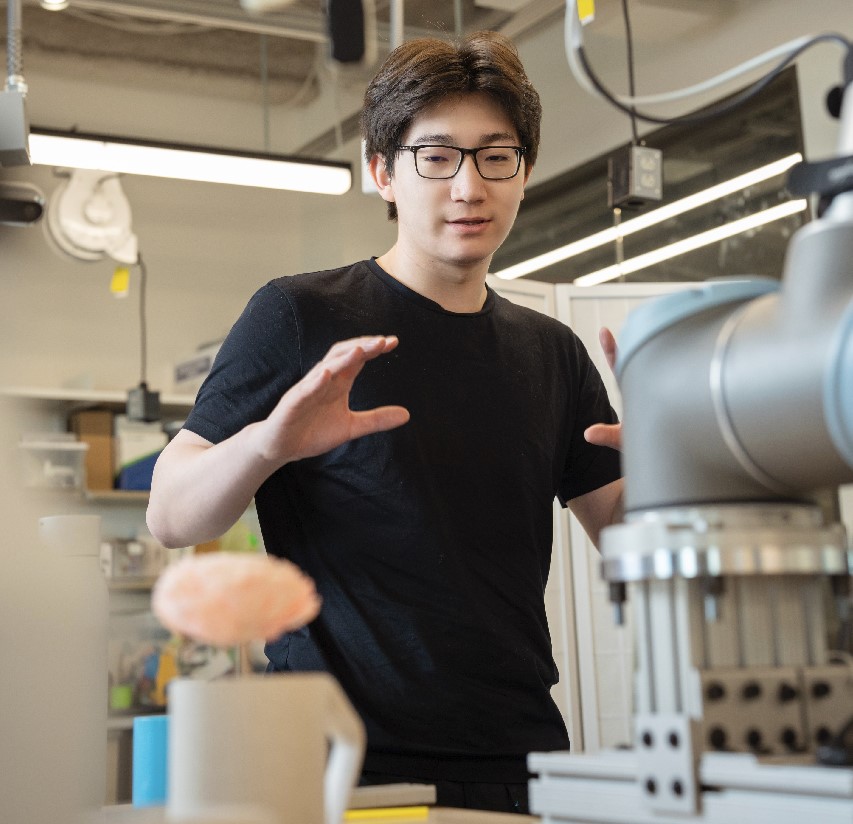}}]{Haotian Liu} is an undergraduate researcher with Professor Ziming Zhang at Worcester Polytechnic Institute and a research intern at Northeastern University with Professor Robert Platt. His research interests include policy learning for robotics, deep learning in 3D vision, FDM printing for soft robotics, and healthcare automation. His works are selected as an oral presentation in RoboSoft 2024 and an oral pitch in IROS 2024.  

\end{IEEEbiography}

\vskip -40pt plus -1fil

\begin{IEEEbiography}[{\includegraphics[width=1in,height=1.25in,clip,keepaspectratio]{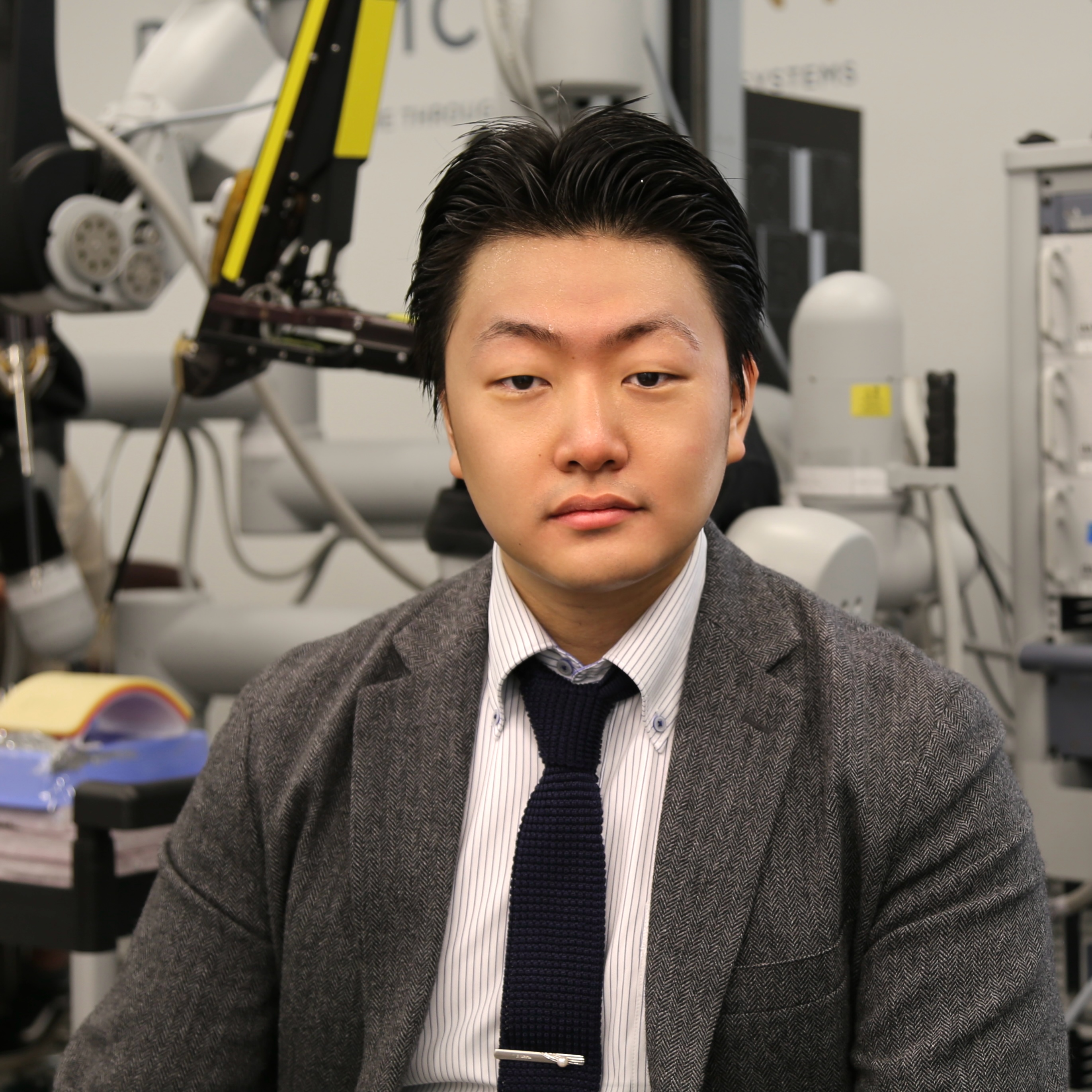}}]{Shang Gao}
born in Nagoya, Japan, is a Ph.D. candidate at the Medical FUSION (Frontier Ultrasound Imaging and Robotic Instrumentation) Laboratory at Worcester Polytechnic Institute (WPI) in the USA. He earned his M.S. degree in Robotics Engineering from WPI in 2020. In 2018, he received dual B.Eng. degrees: Robotics \& Mechatronic Systems Engineering from the University of Detroit Mercy, USA; and Mechanical Engineering from Beijing University of Chemical Technology, China, graduating with honors. His research interests focus on medical robotics, photoacoustic imaging, and image-guided interventions. He is the recipient of the Robert F. Wagner All-Conference Best Student Paper Award from SPIE Medical Imaging 2023.

\end{IEEEbiography}

\vskip -30pt plus -1fil

\begin{IEEEbiography}[{\includegraphics[width=1in,height=1.25in,clip,keepaspectratio]{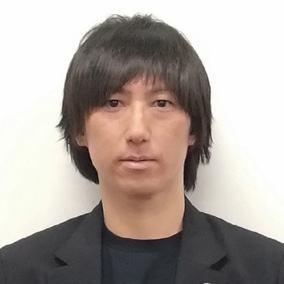}}]{Kazunori D Yamada} is a Professor at Unprecedented-scale Data Analytics Center and Graduate School of Information Sciences of Tohoku University, Japan. He is also a Visiting Researcher at Artificial Intelligence Research Center, National Institute of Advanced Industrial Science and Technology, Japan. His research interest is neural network algorithms for processing context information, and machine consciousness.

\end{IEEEbiography}

\vskip -30pt plus -1fil

\begin{IEEEbiography}[{\includegraphics[width=1in,height=1.25in,clip,keepaspectratio]{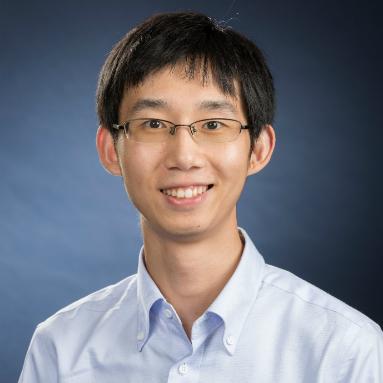}}]{Haichong Zhang} is an Associate Professor at Worcester Polytechnic Institute (WPI). He is the founding director of the Medical Frontier Ultrasound Imaging and Robotic Instrumentation (FUSION) Laboratory. The research in his lab focuses on the interface of medical imaging, sensing, and robotics, developing robotic-assisted imaging systems as well as image-guided robotic interventional platforms. Dr. Zhang received his B.S. and M.S. in Human Health Sciences from Kyoto University, Japan, and subsequently earned his M.S. and Ph.D. in Computer Science from Johns Hopkins University.

\end{IEEEbiography}

\vskip -30pt plus -1fil

\begin{IEEEbiography}[{\includegraphics[width=1in,height=1.25in,clip,keepaspectratio]{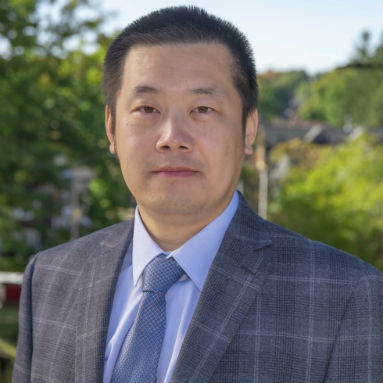}}]{Ziming Zhang} is an assistant professor at Worcester Polytechnic Institute (WPI). Before joining WPI he was a research scientist at Mitsubishi Electric Research Laboratories (MERL) in 2017-2019. Prior to that, he was a research assistant professor at Boston University in 2016-2017. Dr. Zhang received his PhD in 2013 from Oxford Brookes University, UK, under the supervision of Prof. Philip H. S. Torr. His research interests lie in computer vision and machine learning. He won the R\&D 100 Award 2018.
\end{IEEEbiography}




\end{document}